\documentclass[10pt,twocolumn,letterpaper,pagenumbers]{article}

\usepackage{wacv}              % To produce the CAMERA-READY version
\usepackage[accsupp]{axessibility} % Improves PDF readability for those with disabilities.

\usepackage{graphicx}
\usepackage{amsmath}
\usepackage{amssymb}
\usepackage{booktabs}
\usepackage{multirow}
\usepackage{balance}
\usepackage{comment}

\usepackage{xcolor}

\usepackage[pagebackref,breaklinks,colorlinks]{hyperref}

\usepackage[capitalize]{cleveref}
\crefname{section}{Sec.}{Secs.}
\Crefname{section}{Section}{Sections}
\Crefname{table}{Table}{Tables}
\crefname{table}{Tab.}{Tabs.}

\def\confName{WACV}
\def\confYear{2025}

\begin{document}

\title{Self-supervised pre-training with diffusion model for few-shot landmark detection in x-ray images}

\author{
    Roberto Di Via\textsuperscript{1} \quad Francesca Odone\textsuperscript{1} \quad Vito Paolo Pastore\textsuperscript{1}\thanks{Correspondence to: vito.paolo.pastore@unige.it} \\ 
    \textsuperscript{1}MaLGa, DIBRIS, University of Genoa, Italy}
\maketitle

%%%%%%%%% ABSTRACT
\begin{abstract}
Deep neural networks have been extensively applied in the medical domain for various tasks, including image classification, segmentation, and landmark detection. However, their application is often hindered by data scarcity, both in terms of available annotations and images. This study introduces a novel application of denoising diffusion probabilistic models (DDPMs) to the landmark detection task, specifically addressing the challenge of limited annotated data in x-ray imaging.
Our key innovation lies in leveraging DDPMs for self-supervised pre-training in landmark detection, a previously unexplored approach in this domain. This method enables accurate landmark detection with minimal annotated training data (as few as $50$ images), surpassing both ImageNet supervised pre-training and traditional self-supervised techniques across three popular x-ray benchmark datasets.
To our knowledge, this work represents the first application of diffusion models for self-supervised learning in landmark detection, which may offer a valuable pre-training approach in few-shot regimes, for mitigating data scarcity. 
To bolster further development and reproducibility, we provide open access to our code and pre-trained models for a variety of x-ray related applications: \href{https://github.com/Malga-Vision/DiffusionXray-FewShot-LandmarkDetection}{https://github.com/Malga-Vision/DiffusionXray-FewShot-LandmarkDetection}
\end{abstract}

%%%%%%%%% BODY TEXT
\section{Introduction}
\label{sec:intro}
Landmark detection, the task of identifying anatomical keypoints in images \cite{ibragimov2017landmark}, plays a crucial role in various medical applications, including angle measurements \cite{DBLP:conf/isbi/McCouatVG21}, skeletal assessments \cite{DBLP:journals/ijon/LeeCS22}, and surgical planning \cite{edwards2021deepnavnet}. Although deep neural networks have become prevalent in this domain \cite{DBLP:journals/ijmir/SuganyadeviS022,Meijering2020}, and many fully supervised methods have been proposed \cite{Zhu2022,zhu2021you,divia2024indomain,DBLP:conf/miccai/JiangLWTLL22,DBLP:journals/corr/abs-2206-02087,DBLP:conf/miccai/ElkhillLFP22,DBLP:conf/miccai/YaoHHZ20,DBLP:conf/miccai/ViriyasaranonMC23,kasturi2024anatomical}, their effectiveness is often limited by the scarcity of annotated data, particularly in medical imaging where expert annotations are costly and time consuming \cite{zhu2021you,Zhu2022,divia2024indomain}. Consequently, real-world datasets typically contain very few annotated images, requiring the design of label-efficient training solutions \cite{DBLP:conf/miccai/ZhuQYLZ23}.

To address this challenge, researchers have explored transfer learning approaches, typically fine-tuning ImageNet pre-trained models on medical tasks \cite{DBLP:journals/bmcmi/KimCSJMG22,Alzubaidi2021,DBLP:conf/eccv/XieR18}. However, the efficacy of in-domain pre-training remains a subject of debate, with recent studies yielding conflicting results across different tasks and domains \cite{touijer2023food,divia2024indomain}. Self-supervised learning (SSL) methods such as MoCoV3 \cite{DBLP:conf/iccv/ChenXH21}, SimCLRV2 \cite{DBLP:conf/nips/ChenKSNH20}, and DINO \cite{DBLP:conf/iccv/CaronTMJMBJ21} have emerged as promising alternatives, enabling models to learn robust representations from unlabeled data. These approaches have shown potential to reduce the dependency on large labeled datasets, especially for medical imaging applications.

In this context, our primary contribution lies in the novel application of DDPM to the task of landmark detection in x-ray images. While DDPMs have demonstrated remarkable success in image generation tasks \cite{DBLP:journals/corr/abs-2105-05233,DBLP:journals/corr/abs-2212-07501}, their potential for self-supervised pre-training in medical image analysis, particularly landmark detection, remains largely unexplored. We propose a few-shot self-supervised pre-training approach specifically tailored for landmark detection in x-ray images, widely used for anatomical assessment and diagnosis, addressing the realistic and typical scenario where the available annotations and images are minimal (up to $50$). 

To our knowledge, this work represents the first application of diffusion models for self-supervision in landmark detection. We conduct a comprehensive comparison of our method against state-of-the-art alternatives, including YOLO \cite{zhu2021you}, ImageNet supervised pre-training and self-supervised approaches like MoCoV3, SimCLRV2, and DINO. Our evaluation focuses on performance and robustness across varying quantities of labeled training samples.

Our results demonstrate that our DDPM-based approach outperforms both ImageNet supervised pre-training and other self-supervised methods in the landmark detection task. Furthermore, we perform a set of experiments to provide insights on whether the usage of a different in-domain pre-training dataset may benefit the downstream task, in a few shot regimes. 

To facilitate further research in x-ray imaging, we have made our code and pre-trained models (including DDPM, MoCoV3, SimCLRV2, and DINO) publicly available. These resources, trained on the x-ray datasets used in this study, can be accessed at \href{https://github.com/Malga-Vision/DiffusionXray-FewShot-LandmarkDetection}{https://github.com/Malga-Vision/DiffusionXray-FewShot-LandmarkDetection}.

This work not only introduces a novel application of DDPMs for the landmark detection task but also contributes to the ongoing discussion on effective pre-training strategies for medical imaging tasks with limited annotations.

The remainder of the paper is organized as follows: Section \ref{related} reviews related studies, particularly focusing on self-supervised pre-training in the medical domain and landmark detection. Section \ref{approach} describes our proposed approach. Section \ref{experiments} gives an overview of the benchmark datasets employed in this research and explains the evaluation metrics. Lastly, Section \ref{results} showcases our experimental results and insights, drawing some conclusions in Section \ref{conclusion}.

\begin{figure*}[tbh]
    \centering
    \includegraphics[width=1\textwidth]{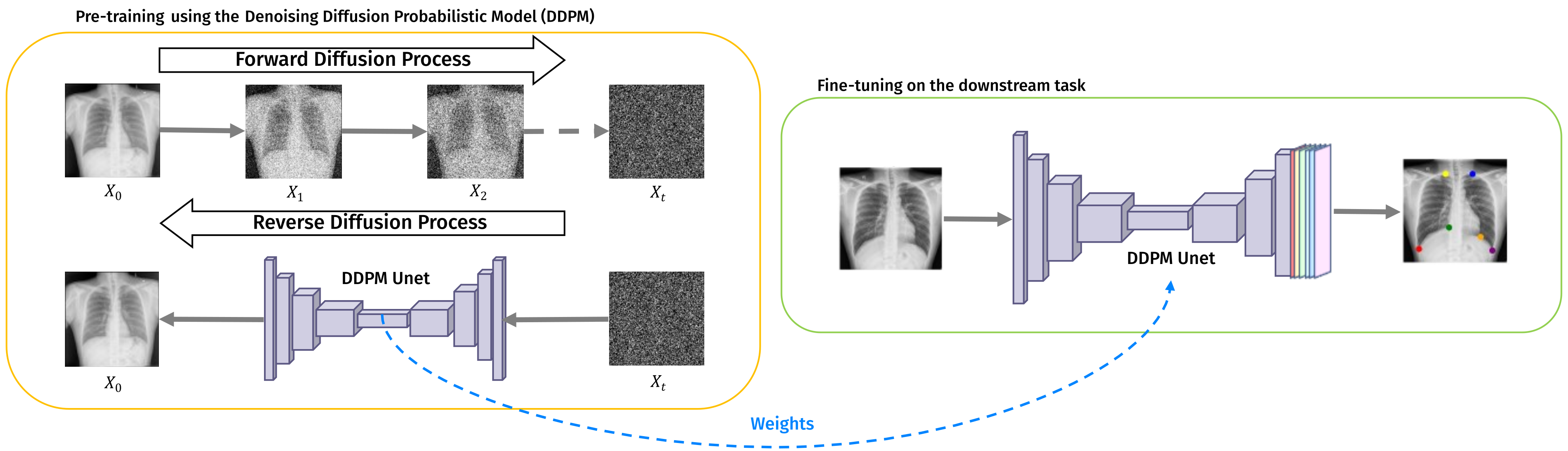}
    \caption{Schematic representation of our DDPM self-supervised landmark detection pipeline.}
    \label{fig:pipeline}
\end{figure*}

\section{Related Works}
\label{related}
Few-shot learning is particularly relevant in medical applications due to the high cost and time-intensive nature of obtaining high-quality, unbiased annotations \cite{DBLP:conf/cvpr/QuanYLZ22}. To address this challenge, researchers have explored various approaches, with transfer learning and self-supervised learning emerging as two prominent strategies.

In landmark detection, existing approaches have primarily used transfer learning to address data scarcity \cite{DBLP:conf/iccvw/TiulpinMS19,zhu2021you,divia2024indomain}. 
Tiulpin et al. \cite{DBLP:conf/iccvw/TiulpinMS19} applied transfer learning from low-budget annotations for knee x-ray landmark localization in osteoarthritis stages. 
Di Via et al. \cite{divia2024indomain} investigated the potential benefits of in-domain data over ImageNet pre-training for x-ray image landmark detection. 
Zhu et al. \cite{zhu2021you} proposed GU2Net, a universal model for anatomical landmark detection in medical images, addressing limitations in dataset-dependent methods.

While transfer learning has shown promise, recent advancements in self-supervised learning have opened new avenues for tackling data scarcity in medical imaging tasks \cite{DBLP:journals/bmcmi/VanBerloHW24}. 
Yao et al. \cite{DBLP:conf/miccai/YaoQXZ21} developed a two-stage framework called Cascade Comparing to Detect (CC2D) that uses self-supervised learning and pseudo-label training.
Zhu et al. \cite{DBLP:conf/miccai/ZhuQYLZ23} introduced UOD, a domain-adaptive framework using contrastive learning and a domain-adaptive transformer. 
Miao et al. \cite{DBLP:journals/corr/abs-2407-05412} proposed FM-OSD, using pre-trained visual foundation models for anatomical landmark detection introducing distance-aware similarity learning loss, and bidirectional matching strategy to enhance detection accuracy.

In parallel with these developments, diffusion models have gained traction in various medical imaging tasks \cite{DBLP:journals/mia/KazerouniAHAFHM23}, although their potential for landmark detection remains largely unexplored.  
Pinaya et al. \cite{DBLP:conf/miccai/PinayaTDCFNOC22} developed a latent diffusion model for for 3D brain MRI generation. 
Rousseau et al. \cite{DBLP:conf/miccai/RousseauACMMA23} used diffusion models for dental radiography segmentation. 
Baranchuk et al. \cite{DBLP:conf/iclr/BaranchukVRKB22} employed diffusion models for semantic segmentation. 
Wyatt et al. \cite{DBLP:conf/cvpr/WyattLSW22} proposed AnoDDPM for anomaly detection in medical images.

Building upon these advancements in both self-supervised learning and diffusion models, our work distinguishes itself by examining the impact of self-supervised pre-training with diffusion models on landmark detection for x-ray images in a realistic scenario with limited annotations and images, a common challenge in medical imaging. 

To our knowledge, this study represents the first application of diffusion models for self-supervised learning in the specific task of landmark detection in x-ray images. This novel approach bridges the gap between the established potential of diffusion models in other medical imaging tasks and the pressing need for label-efficient methods in landmark detection.

%-------------------------------------------------------------------------

\section{Approach}
\label{approach}
Fig. \ref{fig:pipeline} illustrates our two-step methodology: initially, a DDPM is pre-trained on the training set, in a self-supervised fashion (with no annotations). Subsequently, this model is fine-tuned on the labeled training set for the downstream task, represented by the landmark detection. Further details on these steps are discussed below.\\
\\
\textbf{DDPMs background.}
Denoising Diffusion Probabilistic Models (DDPMs) \cite{DBLP:journals/corr/abs-2006-11239} have emerged as a powerful class of generative models, excelling in the synthesis of high-fidelity samples in diverse fields, including medical imaging \cite{DBLP:journals/corr/abs-2105-05233,DBLP:journals/corr/abs-2212-07501}. DDPMs comprise two key processes: forward diffusion and reverse diffusion. \\
\\
1. \textit{Forward Diffusion Process:} \\
The forward diffusion process is described by:
\begin{equation}
q(x_t | x_{t-1}) = \mathcal{N}(x_t; \sqrt{1-\beta_t}x_{t-1}, \beta_t \mathbf{I})
\end{equation}
Here, $q(x_t | x_{t-1})$ represents the probability distribution of $x_t$ given $x_{t-1}$. $\mathcal{N}$ denotes a Gaussian distribution with mean $\sqrt{1-\beta_t}x_{t-1}$ and variance $\beta_t \mathbf{I}$. $\beta_t$ is a variance schedule that increases with $t$, controlling the amount of noise added at each step. $\mathbf{I}$ is the identity matrix. The process can be expressed as a single step:
\begin{equation}
q(x_t | x_0) = \mathcal{N}(x_t; \sqrt{\bar{\alpha}_t}x_0, (1-\bar{\alpha}_t)\mathbf{I})
\end{equation}
where $\bar{\alpha}_t = \prod_{s=1}^t (1-\beta_s)$. This equation allows direct sampling of $x_t$ given $x_0$, with $\sqrt{\bar{\alpha}_t}x_0$ as the mean and $(1-\bar{\alpha}_t)\mathbf{I}$ as the variance. \\
\\ \\ \\
2. \textit{Reverse Diffusion Process:} \\
The reverse diffusion process is modeled as:
\begin{equation}
p_\theta(x_{t-1} | x_t) = \mathcal{N}(x_{t-1}; \mu_\theta(x_t, t), \Sigma_\theta(x_t, t))
\end{equation}
Here, $p_\theta(x_{t-1} | x_t)$ is the probability of $x_{t-1}$ given $x_t$. $\mu_\theta(x_t, t)$ and $\Sigma_\theta(x_t, t)$ are the mean and covariance predicted by a neural network with parameters $\theta$. The mean of the reverse process is estimated by:
\begin{equation}
\mu_\theta(x_t, t) = \frac{1}{\sqrt{\alpha_t}}\left(x_t - \frac{\beta_t}{\sqrt{1-\bar{\alpha}_t}}\epsilon_\theta(x_t, t)\right)
\end{equation}
where $\epsilon_\theta(x_t, t)$ is the noise predicted by the neural network. The terms $\frac{1}{\sqrt{\alpha_t}}$ and $\frac{\beta_t}{\sqrt{1-\bar{\alpha}_t}}$ are scaling factors that adjust for the noise level at time t.\\
\\
The training objective is:
\begin{equation}
\mathcal{L}_{\text{simple}} = \mathbb{E}_{t,x_0,\epsilon} \left[ \|\epsilon - \epsilon_\theta(\sqrt{\bar{\alpha}_t}x_0 + \sqrt{1-\bar{\alpha}_t}\epsilon, t)\|^2 \right]
\end{equation}
This loss function trains the model to predict the noise $\epsilon$ added at each timestep. $\epsilon_\theta(\cdot)$ is the noise predicted by the model, and $\sqrt{\bar{\alpha}_t}x_0 + \sqrt{1-\bar{\alpha}_t}\epsilon$ is the noisy input to the model at time t. The expectation $\mathbb{E}_{t,x_0,\epsilon}$ is taken over random timesteps, images, and noise samples.  \\
\\
\textbf{Pre-training using the DDPM.}
In this work, we leverage a DDPM U-Net architecture as a pre-training strategy for anatomical landmark detection using unannotated medical images. 
This approach captures multi-scale anatomical representations, with DDPM's denoising process extracting features at various levels and U-Net's skip connections preserving both fine details and global information.
During pre-training, the model learns to generate new images with the same pixel distribution as the originals, enabling it to learn important anatomical features.

Given the typical scarcity of labeled data in real-world landmark detection, our DDPM U-Net is designed for effective pre-training on small-scale unannotated datasets while maintaining computational efficiency and minimizing training time. This approach yields rich and general features for the downstream task, offering an effective few-shot learning framework for landmark detection. By leveraging unannotated medical data, our model develops a comprehensive understanding of anatomical structures, creating a robust foundation that can adapt to specific landmark detection tasks with minimal additional training. \\
\\
\textbf{Ground-truth heatmaps generation.}
We formulate landmark detection as a classification problem over all image pixels to generate output heatmaps. This approach enables simultaneous prediction of multiple landmarks, with one heatmap per landmark. Specifically, our model outputs $N$ heatmaps, where $N$ is the number of landmarks to predict.
For heatmap generation, we adopt a Gaussian-based approach inspired by \cite{zhu2021you}. This strategy involves applying a Gaussian filter ($\sigma = 5$) centered at the ground-truth landmark location, followed by thresholding to obtain binary images. Formally:
\[
H_g(i,j) = \begin{cases}
1, & \text{if } \exp\left(-\frac{(i-x)^2 + (j-y)^2}{2\sigma^2}\right) > \frac{1}{2}\max(G) \\
0, & \text{otherwise}
\end{cases}
\]
Where $H_g(i,j)$ is the Gaussian heatmap pixel value at $(i,j)$, $(x,y)$ are the ground truth landmark coordinates, and $G$ is the Gaussian-filtered image before thresholding. \\
\\
Training on these heatmaps enables our model to learn landmark spatial relationships, predicting both general areas of interest and precise coordinates, and potentially enhance downstream landmark detection.\\
\\ 
\textbf{Fine-tuning on the downstream task.} 
The pre-trained DDPM UNet model is adapted for anatomical landmark detection, bridging the gap between generative modeling and precise anatomical localization. This adaptation, illustrated in Figure \ref{fig:pipeline}, involves a subtle yet powerful modification of the UNet architecture. While preserving the core structure that proved so effective in the DDPM, we alter the final convolutional layer to output multiple channels (each corresponding to a predicted heatmap for a specific anatomical landmark) and the timestep embedding is set to be null (as the temporal aspect is irrelevant for static landmark detection). This architectural shift transforms the model from a general image generator to a specialized landmark detector, capitalizing on learned hierarchical representations.

Our fine-tuning strategy transitions from DDPM's self-supervised paradigm to a supervised approach tailored for landmark detection. Training with ground-truth heatmaps allows the model to repurpose its learned features for precise spatial localization. Landmark coordinates are derived by calculating the centroid of each heatmap generated.

This supervised fine-tuning, combined with pre-trained weight transfer (indicated by the blue dashed arrow), enables quick adaptation to the new task with minimal annotated data. The approach's efficacy in few-shot learning contexts highlights its potential for medical imaging applications, where large annotated datasets are often scarce. By harnessing the power of diffusion models, traditionally used for generation tasks, we show their unexplored potential in discriminative tasks like landmark detection.

%-------------------------------------------------------------------------
\section{Experimental Setup}
\label{experiments}
\subsection{Datasets and Evaluation Metrics}
\label{sec:dataset}
We evaluate our method on three public x-ray datasets standard in landmark detection research, following the dataset splits and evaluation protocols of \cite{zhu2021you,divia2024indomain}. The \textbf{Chest x-ray dataset} \cite{DBLP:journals/tmi/CandemirJPMSXKATM14} comprises $279$ images (approximately $3000$×$3000$ pixels, resolution unknown) with $6$ landmarks, split into $195$ for training, $34$ for validation, and $50$ for testing. The \textbf{Cephalometric x-ray dataset} \cite{DBLP:journals/mia/WangHLLCSLIVRFC16} consists of $400$ lateral cephalograms ($2400$×$1935$ pixels, $0.1$ mm resolution) with $19$ target landmarks, split into $130$ for training, $20$ for validation, and $250$ for testing. The \textbf{Hand x-ray dataset} \cite{DBLP:journals/cmig/GertychZSPH07} contains $909$ radiographs (average size $1563$×$2169$ pixels) with $37$ labeled landmarks, assuming a $50$ mm length between wrist endpoints \cite{DBLP:journals/mia/PayerSBU19}, split into $550$ for training, $59$ for validation, and $300$ for testing. 

Our model is assessed using two standard metrics \cite{DBLP:conf/cvpr/QuanYLZ22,DBLP:conf/eccv/YinGWYW22}: the \textit{mean radial error} (MRE) and the \textit{successful detection rate} (SDR). MRE measures the Euclidean distance between predicted and ground truth landmarks, while SDR calculates the percentage of predictions within dataset-specific thresholds. For the Chest dataset, we report these metrics in pixels due to unavailable physical spacing information, whereas for Cephalometric and Hand datasets, we use millimeters, adhering to field standards.

%-------------------------------------------------------------------------

\subsection{Implementation Details} 
\label{sec:details}
%This section provides the technical specifics of the implementation and experimental setup. 
The experiments are conducted on a single NVIDIA A30 GPU with 24 GB RAM, using Python 3.10 and PyTorch 2.1.0. For data pre-processing, images are resized to $256$x$256$ pixels, normalized to $[0-1]$ range, and augmented during training with random rotations ($-2$\textdegree and $2$\textdegree), scaling ($-0.02$ to $0.02$), and translations ($-0.02$ to $0.02$).\\
\\
\textbf{Pre-training stage.} The DDPM is trained with a batch size of $4$, with gradient accumulation every $8$ batches. Training involves $500$ diffusion steps with a linear noise schedule. The training process spans $10k$ iterations, and we save the weights at iterations $4k$, $6k$, $8k$, and $10k$ for comparison during fine-tuning. We utilize the AdamW optimizer with a learning rate of $1e^{-4}$ and apply the exponential moving average to model parameters with a decay factor of $0.995$. We employ mean squared error as the loss function, consistent with the paper \cite{DBLP:journals/corr/abs-2006-11239}. 

The DDPM Unet architecture features a multi-scale approach with channel multipliers [$1$, $2$, $4$, $8$], indicating the increase in channels at each down-sampling stage. It incorporates attention mechanisms at a resolution of $32$, uses $4$ attention heads per channel, and includes $4$ residual blocks at each resolution level. 

Concerning the implementation of other SSL and ImageNet pre-training techniques, the sources are \cite{pyssl2023giakoumoglou} and \cite{Iakubovskii:2019}. \\ 
\\
\textbf{Fine-tuning stage.} The fine-tuning process is carried out using the AdamW optimizer for $200$ epochs, with an early stopping criterion. The learning rate is adjusted using a ReduceLROnPlateau scheduler based on the validation loss, starting at an initial value of $1e^{-5}$. The batch size is set at $2$, with a gradient accumulation of $8$. We use cross-entropy loss for gaussian heatmaps, and a negative log-likelihood loss for the contour-hugging ones \cite{DBLP:conf/cvpr/McCouatV22}.

%-------------------------------------------------------------------------

\section{Experimental Results}
\label{results}

\subsection{Tuning the DDPM pre-training iterations}
\label{sec:impact_iterations}
\begin{figure}[thb]
    \centering
    \includegraphics[width=0.85\linewidth]{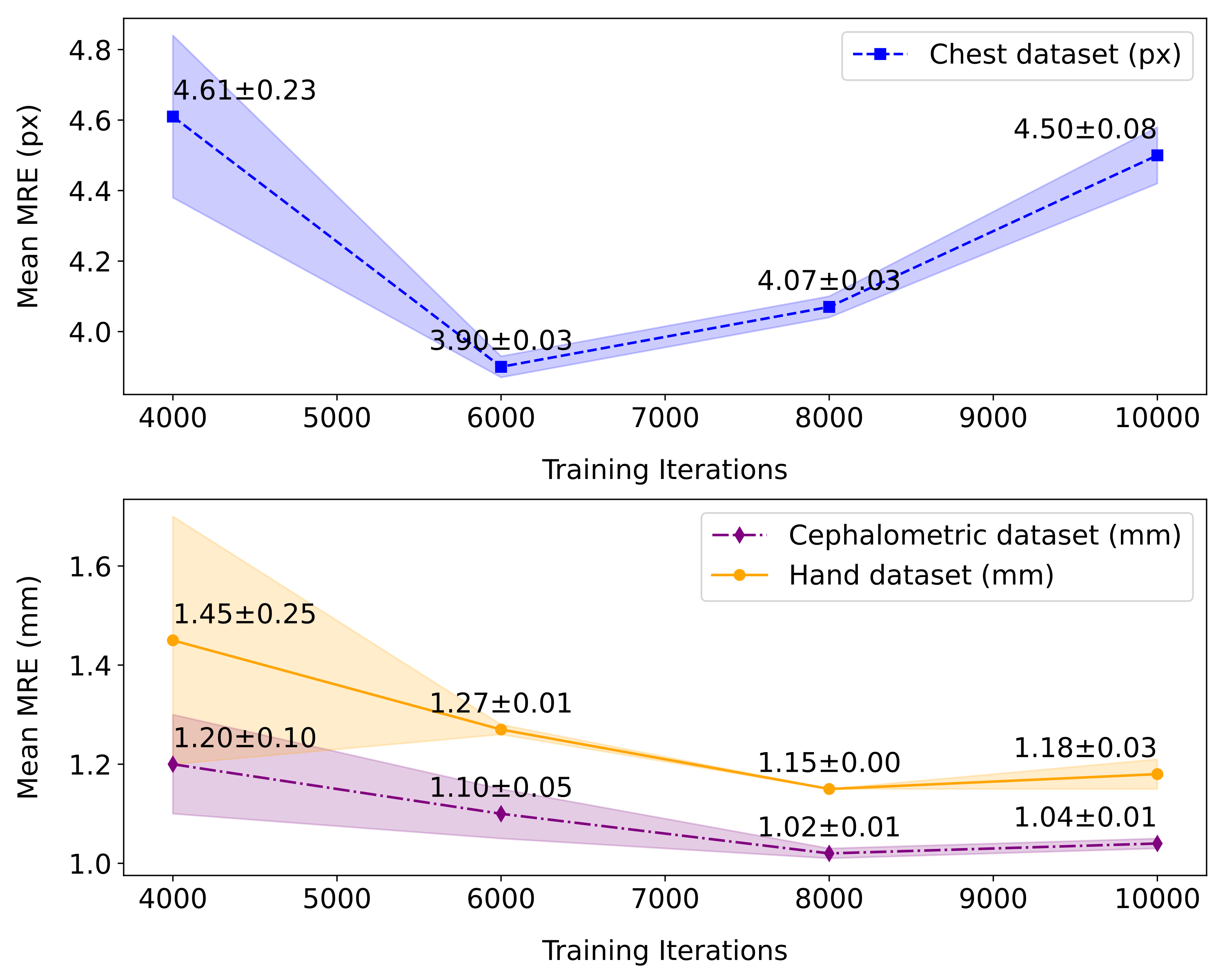}
    \caption{Landmark detection performance on Chest, Cephalometric, and Hand validation sets across DDPM pre-training iterations.}
    \label{fig:ddpm_tuning}
\end{figure}

The number of training iterations during DDPM pre-training is a fundamental hyperparameter in our approach. To evaluate its impact and determine the optimal value, we conducted an experiment running the entire pipeline three times, considering $4k$, $6k$, $8k$, and $10k$ iterations for the DDPM pre-training step. 
Figure \ref{fig:ddpm_tuning} illustrates the landmark detection results obtained on the validation sets for the Chest, Cephalometric, and Hand datasets in terms of MRE.

Our findings indicate that the pre-training is quite robust, except for the $4k$ iterations where the model is likely still learning crucial data features (as shown by the high standard deviation).  Excessive training iterations beyond $8k$ may result in overfitting, resulting in slightly degraded performance on the downstream task. Based on these results, we selected the optimal number of training iterations for DDPM pre-training in subsequent experiments, utilizing $6k$, $8k$, and $8k$ iterations for the Chest, Cephalometric, and Hand datasets respectively.

\subsection{Downstream task performance evaluation}
We assess the effectiveness of our DDPM self-supervised pre-training method by benchmarking it against supervised ImageNet pre-training and self-supervised state-of-the-art methods (MoCoV3, SimCLRV2, and DINO) across different numbers of labeled training samples (1, 5, 10, 25, and 50) in the Chest, Cephalometric, and Hand datasets. Additionally, we explore an ablation of our method using the contour-hugging approach \cite{DBLP:conf/cvpr/McCouatV22} as an alternative framework for the downstream task. Tables \ref{tab:downstream_chest}, \ref{tab:downstream_cephalometric}, and \ref{tab:downstream_hand} report the average test results for three independent runs (a bar chart version of the tables is provided in the Supplementary Material). 
Except for our approach, which implements a DDPM as described in \cite{DBLP:journals/corr/abs-2006-11239}, all methods use DenseNet161 as the encoder backbone, chosen for its performance across all datasets based on hold-out validation comparisons with VGG19 and ResNeXt50\_32x4D.

\begin{figure*}[!ht]
    \centering
    \includegraphics[width=0.9\textwidth]{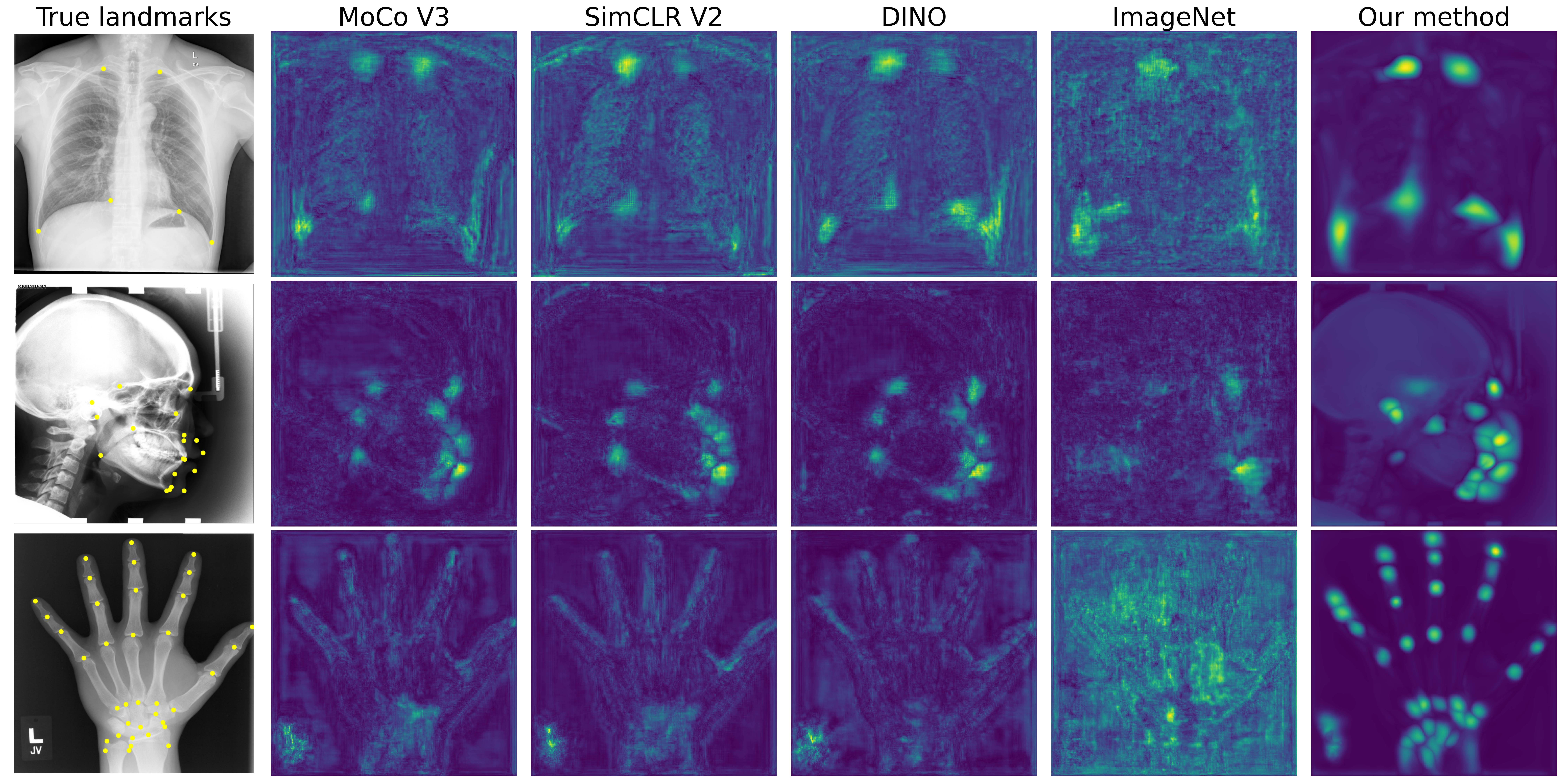}
    \caption{Comparison of landmark detection performance between our DDPM pre-training method and alternative self-supervised and ImageNet pre-training approaches on Chest, Cephalometric, and Hand x-ray test sets, using only 10 labeled training samples.}
    \label{fig:heatmaps}
\end{figure*}

Our approach consistently outperforms both ImageNet and alternative SSL approaches across all datasets and training image quantities. Figure \ref{fig:heatmaps} provides visual support, showing prediction heatmaps for each method with 10 labels across all datasets. Performance gains are particularly pronounced in low-data regimes. For instance, in the Chest dataset (Table \ref{tab:downstream_chest}) with one labeled sample, our method achieves an MRE of $14.99$px compared to $143.67$px for ImageNet and $64.69$px for DINO (the best alternative), representing reductions of $89.6$\% and $76.8$\% respectively. The SDR at $9$px shows even greater improvements, with our method reaching $64.63$\%, far surpassing other approaches.

\begin{table*}[!ht]
    \centering
    \caption{Comparing landmark detection test performance on the Chest dataset through our DDPM self-supervised pre-training and ImageNet, MoCoV3, SimCLRV2, DINO pre-training across varying numbers of labeled training images, highlighting best results in bold.}
    \label{tab:downstream_chest}
    {\small {
    \resizebox{0.7\textwidth}{!}{%
    
        \begin{tabular}{ c l | c c c c  }
        \toprule
             \multirow{3}{*}{\shortstack{Number of Labeled \\Training Samples}} & 
             \multirow{3}{*}{\shortstack{Pre-trained \\Models}} & 
             \multicolumn{4}{c}{\textbf{Chest}} \\
             & & MRE $\downarrow$ & \multicolumn{3}{c}{SDR(\%) $\uparrow$} \\
            \cline{4-6}
            & & (px) & 3px & 6px & 9px \\

        \toprule
        \multirow{5}{*}{$1$}                                
        & ImageNet    & $143.67\pm0.27$ & $0.00\pm0.00$ & $0.00\pm0.00$ & $0.00\pm0.00$ \\
                                
        & MoCoV3      & $106.74\pm14.95$ & $2.98\pm0.51$ & $12.33\pm2.26$ & $19.92\pm3.47$ \\

        & SimCLRV2    & $118.97\pm1.84$ & $1.90\pm0.19$ & $6.09\pm0.58$ & $11.39\pm0.58$ \\

        & DINO        & $64.69\pm0.91$ & $3.66\pm0.33$ & $15.72\pm0.38$ & $23.44\pm0.38$ \\

        & Our method with \cite{DBLP:conf/cvpr/McCouatV22} & $50.03\pm0.00$ & $\mathbf{21.14\pm0.00}$ & $43.90\pm0.00$ & $50.81\pm0.00$ \\
        & Our method       & $\mathbf{14.99\pm0.00}$ & $19.92\pm0.00$ & $\mathbf{46.34\pm0.00}$ & $\mathbf{64.63\pm0.00}$ \\

        \midrule
        
        \multirow{5}{*}{$5$}                                
        & ImageNet    & $72.81\pm2.71$ & $2.44\pm0.33$ & $8.54\pm1.00$ & $14.09\pm1.38$ \\
                                
        & MoCoV3      & $18.35\pm0.18$ & $14.63\pm1.00$ & $41.60\pm0.69$ & $66.53\pm0.69$ \\

        & SimCLRV2    & $30.09\pm0.35$ & $15.72\pm0.69$ & $48.37\pm1.45$ & $67.48\pm0.33$ \\

        & DINO        & $24.48\pm1.73$ & $11.25\pm1.16$ & $32.65\pm0.51$ & $48.78\pm1.00$ \\
        
        & Our method with \cite{DBLP:conf/cvpr/McCouatV22} & $8.14\pm0.00$ & $\mathbf{60.16\pm0.00}$ & $\mathbf{82.11\pm0.00}$ & $\mathbf{86.99\pm0.00}$ \\

        & Our method       & $\mathbf{6.04\pm0.00}$ & $32.52\pm0.00$ & $65.04\pm0.00$ & $\mathbf{86.99\pm0.00}$ \\

        \midrule
        
        \multirow{5}{*}{$10$}
        
        & ImageNet    & $35.83\pm1.57$ & $7.11\pm0.84$ & $22.56\pm1.06$ & $37.30\pm1.39$ \\
                                
        & MoCoV3      & $13.52\pm0.67$ & $18.49\pm0.21$ & $44.72\pm1.21$ & $67.28\pm1.43$ \\

        & SimCLRV2    & $28.50\pm0.00$ & $17.89\pm0.00$ & $45.93\pm0.00$ & $64.63\pm0.00$ \\

        & DINO        & $14.96\pm0.03$ & $16.26\pm0.41$ & $39.64\pm0.21$ & $61.99\pm0.61$ \\
        
        & Our method with \cite{DBLP:conf/cvpr/McCouatV22} & $\mathbf{4.47\pm0.00}$ & $\mathbf{60.98\pm0.00}$ & $\mathbf{88.62\pm0.00}$ & $\mathbf{93.50\pm0.00}$ \\

        & Our method          & $5.33\pm0.00$ & $33.74\pm0.00$ & $67.89\pm0.00$ & $86.59\pm0.00$ \\

        \midrule
        \multirow{5}{*}{$25$}                                
        & ImageNet    & $11.38\pm1.11$ & $20.83\pm2.00$ & $49.29\pm1.13$ & $70.33\pm0.29$ \\
                                
        & MoCoV3      & $6.15\pm0.00$ & $30.89\pm0.00$ & $66.67\pm0.00$ & $85.37\pm0.00$ \\

        & SimCLRV2    & $11.46\pm0.00$ & $33.74\pm0.00$ & $69.11\pm0.00$ & $84.15\pm0.00$ \\

        & DINO        & $10.32\pm0.35$ & $24.59\pm1.02$ & $55.69\pm0.81$ & $75.81\pm0.20$ \\       

        & Our method with \cite{DBLP:conf/cvpr/McCouatV22} & $\mathbf{2.92\pm0.00}$ & $\mathbf{76.02\pm0.00}$ & $\mathbf{94.31\pm0.00}$ & $\mathbf{96.34\pm0.00}$ \\

        & Our method          & $4.62\pm0.00$ & $32.93\pm0.00$ & $73.17\pm0.00$ & $91.87\pm0.00$ \\

        \midrule
        \multirow{5}{*}{$50$}
        
        & ImageNet    & $7.36\pm0.31$ & $29.98\pm1.73$ & $67.58\pm2.45$ & $83.23\pm0.34$ \\
                                
        & MoCoV3      & $4.42\pm0.00$ & $39.43\pm0.00$ & $77.64\pm0.00$ & $92.68\pm0.00$ \\

        & SimCLRV2    & $7.93\pm0.00$ & $37.40\pm0.00$ & $73.98\pm0.00$ & $91.87\pm0.00$ \\

        & DINO        & $7.62\pm1.04$ & $30.89\pm1.22$ & $62.81\pm1.43$ & $82.32\pm3.05$ \\      

        & Our method with \cite{DBLP:conf/cvpr/McCouatV22} & $\mathbf{2.69\pm0.00}$ & $\mathbf{76.83\pm0.00}$ & $\mathbf{93.50\pm0.00}$ & $\mathbf{97.56\pm0.00}$ \\

        & Our method          & $4.31\pm0.01$ & $39.63\pm0.61$ & $80.08\pm0.41$ & $93.50\pm0.41$ \\

        \bottomrule
        \end{tabular}}}
    }
\end{table*}

Similar trends are observed in the Cephalometric (Table \ref{tab:downstream_cephalometric}) and Hand datasets (Table \ref{tab:downstream_hand}). With one labeled sample in the Cephalometric dataset, our method attains an MRE of $15.71$mm versus $86.71$mm for ImageNet and $43.49$mm for MoCoV3, with corresponding improvements in SDR. In the Hand dataset, our approach yields an MRE of $28.75$mm versus $79.32$mm for ImageNet and $97.31$mm for DINO, again with SDR improvements. As the number of labeled samples increases, the performance gap narrows, though our approach maintains an edge. For example, in the Cephalometric dataset with $25$ labeled samples, our method's MRE of $2.84$mm still outperforms ImageNet's $7.67$mm and SimCLRV2's $3.99$mm. This trend is consistent across all datasets, indicating that our method provides substantial benefits in extremely low-data scenarios while continuing to offer improvements as more labeled data becomes available. 
These results highlight our DDPM self-supervised pre-training's effectiveness for landmark detection in few-shot learning scenarios, common in medical imaging where high-quality annotated images are limited.

\begin{table*}[tbhp]
    \centering
    \caption{Comparing landmark detection test performance on Cephalometric dataset through our DDPM self-supervised pre-training and ImageNet, MoCoV3, SimCLRV2, DINO pre-training across varying numbers of labeled training images, highlighting best results in bold.}
    \label{tab:downstream_cephalometric}
    {\small {
    \resizebox{0.691\textwidth}{!}{%
    
        \begin{tabular}{ c l | c c c c c }
        \toprule
             \multirow{3}{*}{\shortstack{Number of Labeled \\Training Samples}} & 
             \multirow{3}{*}{\shortstack{Pre-trained \\Models}} & 
             \multicolumn{5}{c}{\textbf{Cephalometric}} \\
             & & MRE $\downarrow$ & \multicolumn{4}{c}{SDR(\%) $\uparrow$} \\
            \cline{4-7}
            & & (mm) & 2mm & 2.5mm & 3mm & 4mm \\

        \toprule
        \multirow{5}{*}{$1$}                                
        & ImageNet    & $86.71\pm0.02$ & $0.02\pm0.00$ & $0.06\pm0.00$ & $0.06\pm0.00$ & $0.15\pm0.00$ \\
                                
        & MoCoV3      & $43.49\pm1.21$ & $6.79\pm0.10$ & $10.83\pm0.25$ & $13.67\pm0.30$ & $19.99\pm0.30$ \\

        & SimCLRV2    & $100.68\pm12.18$ & $0.82\pm0.42$ & $1.27\pm0.68$ & $1.63\pm0.79$ & $2.51\pm1.19$ \\

        & DINO        & $58.67\pm1.22$ & $2.47\pm0.14$ & $4.42\pm0.07$ & $5.52\pm0.07$ & $8.60\pm0.30$ \\

        & Our method with \cite{DBLP:conf/cvpr/McCouatV22} & $20.13\pm0.00$ & $\mathbf{18.06\pm0.00}$ & $25.14\pm0.00$ & $31.94\pm0.00$ & $41.37\pm0.00$ \\

        & Our method       & $\mathbf{15.71\pm0.00}$ & $17.31\pm0.00$ & $\mathbf{27.14\pm0.00}$ & $\mathbf{33.24\pm0.00}$ & $\mathbf{45.14\pm0.00}$ \\

        \midrule

        \multirow{5}{*}{$5$}                                
        & ImageNet    & $84.40\pm0.02$ & $0.06\pm0.00$ & $0.11\pm0.00$ & $0.15\pm0.00$ & $0.17\pm0.00$ \\
                                
        & MoCoV3      & $13.44\pm0.12$ & $18.62\pm0.19$ & $29.56\pm0.18$ & $35.99\pm0.02$ & $49.38\pm0.16$ \\

        & SimCLRV2    & $24.59\pm0.12$ & $13.06\pm0.31$ & $21.24\pm0.25$ & $26.44\pm0.23$ & $36.52\pm0.37$ \\

        & DINO        & $23.83\pm0.21$ & $14.00\pm0.26$ & $23.01\pm0.54$ & $28.42\pm0.52$ & $39.13\pm0.38$ \\

        & Our method with \cite{DBLP:conf/cvpr/McCouatV22} & $11.53\pm0.00$ & $\mathbf{45.05\pm0.00}$ & $\mathbf{54.00\pm0.00}$ & $\mathbf{61.03\pm0.00}$ & $\mathbf{70.55\pm0.00}$ \\

        & Our method       & $\mathbf{8.30\pm0.00}$ & $27.18\pm0.00$ & $42.65\pm0.00$ & $51.01\pm0.00$ & $66.34\pm0.00$ \\

        \midrule
        
        \multirow{5}{*}{$10$}
                                
        & ImageNet    & $50.09\pm0.44$ & $3.18\pm0.16$ & $5.92\pm0.25$ & $7.49\pm0.26$ & $11.19\pm0.42$ \\
                                
        & MoCoV3      & $9.30\pm0.03$ & $23.84\pm0.12$ & $37.20\pm0.13$ & $45.34\pm0.17$ & $60.18\pm0.13$ \\

        & SimCLRV2    & $18.54\pm0.09$ & $17.80\pm0.29$ & $28.16\pm0.04$ & $34.65\pm0.22$ & $46.74\pm0.23$ \\

        & DINO        & $15.64\pm0.26$ & $18.86\pm0.12$ & $30.89\pm0.10$ & $38.22\pm0.34$ & $51.83\pm0.45$ \\

        & Our method with \cite{DBLP:conf/cvpr/McCouatV22} & $5.21\pm0.00$ & $\mathbf{52.11\pm0.00}$ & $\mathbf{62.97\pm0.00}$ & $\mathbf{71.43\pm0.00}$ & $\mathbf{81.68\pm0.00}$ \\

        & Our method          & $\mathbf{3.52\pm0.00}$ & $32.06\pm0.00$ & $48.59\pm0.00$ & $58.97\pm0.00$ & $74.69\pm0.00$ \\

        \midrule
        \multirow{5}{*}{$25$}
        
        & ImageNet    & $7.67\pm0.39$ & $24.25\pm0.10$ & $37.83\pm0.13$ & $45.58\pm0.26$ & $62.56\pm0.35$ \\
                                
        & MoCoV3      & $4.04\pm0.03$ & $34.81\pm0.18$ & $51.26\pm0.14$ & $60.80\pm0.03$ & $76.78\pm0.10$ \\

        & SimCLRV2    & $3.99\pm0.04$ & $30.63\pm0.23$ & $47.04\pm0.18$ & $56.20\pm0.13$ & $73.29\pm0.21$ \\

        & DINO        & $5.38\pm0.19$ & $30.00\pm0.20$ & $45.98\pm0.04$ & $55.36\pm0.06$ & $72.30\pm0.19$ \\

        & Our method with \cite{DBLP:conf/cvpr/McCouatV22} & $\mathbf{2.72\pm0.00}$ & $\mathbf{65.47\pm0.00}$ & $\mathbf{75.77\pm0.00}$ & $\mathbf{82.55\pm0.00}$ & $\mathbf{90.61\pm0.00}$ \\

        & Our method          & $2.84\pm0.00$ & $39.84\pm0.01$ & $57.61\pm0.01$ & $67.17\pm0.01$ & $81.01\pm0.00$ \\

        \midrule
        \multirow{5}{*}{$50$}
                                
        & ImageNet    & $2.82\pm0.09$ & $42.93\pm2.66$ & $60.74\pm3.48$ & $69.69\pm3.28$ & $83.87\pm2.24$ \\
                                
        & MoCoV3      & $2.98\pm0.01$ & $43.68\pm0.32$ & $61.84\pm0.13$ & $70.48\pm0.03$ & $84.81\pm0.19$ \\

        & SimCLRV2    & $3.17\pm0.04$ & $39.94\pm0.19$ & $58.70\pm0.08$ & $67.66\pm0.05$ & $81.85\pm0.10$ \\

        & DINO        & $2.98\pm0.02$ & $40.33\pm0.38$ & $59.06\pm0.39$ & $68.99\pm0.42$ & $83.78\pm0.64$ \\

        & Our method with \cite{DBLP:conf/cvpr/McCouatV22} & $\mathbf{2.13\pm0.00}$ & $\mathbf{73.60\pm0.00}$ & $\mathbf{81.22\pm0.00}$ & $\mathbf{86.78\pm0.00}$ & $\mathbf{92.70\pm0.00}$ \\

        & Our method          & $2.50\pm0.00$ & $44.91\pm0.05$ & $62.84\pm0.02$ & $72.82\pm0.00$ & $85.81\pm0.02$ \\

        \bottomrule
        \end{tabular} }}
    }
\end{table*}

\begin{table*}[tbh]
    \centering
    \caption{Comparing landmark detection test performance on the Hand dataset through our DDPM self-supervised pre-training and ImageNet, MoCoV3, SimCLRV2, and DINO pre-training across varying numbers of labeled training images, highlighting best results in bold.}
    \label{tab:downstream_hand}
    {\small {
    \resizebox{0.63\textwidth}{!}{%
    
        \begin{tabular}{ c l | c c c c  }
        \toprule
             \multirow{3}{*}{\shortstack{Number of Labeled \\Training Samples}} & 
             \multirow{3}{*}{\shortstack{Pre-trained \\Models}} & 
             \multicolumn{4}{c}{\textbf{Hand}} \\
             & & MRE $\downarrow$ & \multicolumn{3}{c}{SDR(\%) $\uparrow$} \\
            \cline{4-6}
            & & (mm) & 2mm & 4mm & 10mm \\

        \toprule
        \multirow{5}{*}{$1$}
                                
        & ImageNet    & $79.32\pm0.03$ & $0.04\pm0.00$ & $0.14\pm0.00$ & $1.08\pm0.02$ \\
                                
        & MoCoV3      & $100.17\pm0.10$ & $0.42\pm0.02$ & $0.83\pm0.02$ & $1.60\pm0.03$ \\

        & SimCLRV2    & $98.52\pm0.09$ & $0.06\pm0.01$ & $0.31\pm0.00$ & $1.02\pm0.04$ \\

        & DINO        & $97.31\pm0.09$ & $0.47\pm0.04$ & $1.27\pm0.03$ & $3.30\pm0.18$ \\

        & Our method with \cite{DBLP:conf/cvpr/McCouatV22} & $47.76\pm0.00$ & $8.70\pm0.00$ & $12.76\pm0.00$ & $16.59\pm0.00$ \\

        & Our method       & $\mathbf{28.75\pm0.00}$ & $\mathbf{27.87\pm0.00}$ & $\mathbf{46.44\pm0.00}$ & $\mathbf{58.75\pm0.00}$ \\
        
        \midrule

        \multirow{5}{*}{$5$}    
        & ImageNet    & $86.12\pm0.03$ & $0.01\pm0.01$ & $0.09\pm0.01$ & $1.01\pm0.01$ \\
                                
        & MoCoV3      & $61.20\pm0.39$ & $9.82\pm0.16$ & $18.94\pm0.29$ & $32.11\pm0.61$ \\

        & SimCLRV2    & $75.07\pm0.48$ & $5.99\pm0.07$ & $11.87\pm0.14$ & $20.35\pm0.34$ \\

        & DINO        & $86.03\pm0.64$ & $2.05\pm0.31$ & $5.79\pm0.55$ & $11.96\pm0.84$ \\

        & Our method with \cite{DBLP:conf/cvpr/McCouatV22} & $37.53\pm0.00$ & $21.49\pm0.00$ & $28.90\pm0.00$ & $34.17\pm0.00$ \\

        & Our method       & $\mathbf{6.86\pm0.00}$ & $\mathbf{61.09\pm0.00}$ & $\mathbf{83.30\pm0.00}$ & $\mathbf{90.62\pm0.00}$ \\
        
        \midrule
        
        \multirow{5}{*}{$10$}   
        & ImageNet    & $85.87\pm0.00$ & $0.01\pm0.00$ & $0.15\pm0.00$ & $0.99\pm0.00$ \\
                                
        & MoCoV3      & $55.43\pm0.29$ & $13.92\pm0.17$ & $26.00\pm0.15$ & $40.21\pm0.20$ \\

        & SimCLRV2    & $55.58\pm0.75$ & $16.72\pm0.36$ & $29.45\pm0.45$ & $42.18\pm0.72$ \\

        & DINO        & $60.58\pm0.53$ & $11.09\pm0.14$ & $23.00\pm0.12$ & $35.47\pm0.32$ \\

        & Our method with \cite{DBLP:conf/cvpr/McCouatV22} & $25.47\pm0.00$ & $35.67\pm0.00$ & $45.34\pm0.00$ & $51.01\pm0.00$ \\

        & Our method       & $\mathbf{4.94\pm0.11}$ & $\mathbf{62.55\pm0.55}$ & $\mathbf{86.08\pm0.25}$ & $\mathbf{94.28\pm0.17}$ \\

        \midrule
        \multirow{5}{*}{$25$}              
        & ImageNet    & $24.96\pm0.47$ & $42.79\pm0.60$ & $57.93\pm0.28$ & $71.68\pm0.18$ \\
                                
        & MoCoV3      & $40.03\pm0.12$ & $27.19\pm0.09$ & $44.30\pm0.27$ & $56.55\pm0.05$ \\

        & SimCLRV2    & $40.02\pm0.11$ & $29.69\pm0.06$ & $47.03\pm0.00$ & $59.73\pm0.11$ \\

        & DINO        & $49.27\pm0.30$ & $21.09\pm0.06$ & $36.12\pm0.10$ & $48.76\pm0.20$ \\

        & Our method with \cite{DBLP:conf/cvpr/McCouatV22} & $11.99\pm0.00$ & $62.21\pm0.00$ & $77.03\pm0.00$ & $81.39\pm0.00$ \\

        & Our method       & $\mathbf{2.39\pm0.00}$ & $\mathbf{72.99\pm0.00}$ & $\mathbf{93.18\pm0.00}$ & $\mathbf{97.87\pm0.00}$ \\

        \midrule
        \multirow{5}{*}{$50$}                                  
        & ImageNet    & $10.79\pm0.23$ & $61.98\pm0.36$ & $78.51\pm0.26$ & $90.04\pm0.28$ \\
                                
        & MoCoV3      & $23.77\pm0.34$ & $47.44\pm0.10$ & $67.96\pm0.14$ & $76.18\pm0.23$ \\

        & SimCLRV2    & $23.95\pm0.18$ & $49.28\pm0.30$ & $68.56\pm0.45$ & $77.03\pm0.69$ \\

        & DINO        & $34.95\pm0.59$ & $34.06\pm0.97$ & $52.32\pm1.20$ & $64.21\pm0.80$ \\  

        & Our method with \cite{DBLP:conf/cvpr/McCouatV22} & $7.16\pm0.00$ & $\mathbf{79.38\pm0.00}$ & $88.36\pm0.00$ & $90.19\pm0.00$ \\

        & Our method          & $\mathbf{1.74\pm0.03}$ & $78.04\pm0.94$ & $\mathbf{96.35\pm0.51}$ & $\mathbf{99.02\pm0.04}$ \\

        \bottomrule
        \end{tabular} }}
    }
  
\end{table*}

 \begin{table*}[!ht]
   \centering
   \caption{Comparison of our method vs. state-of-the-art YOLO framework using 1, 5, and 10 labeled samples. + indicates mixed dataset training. Both methods utilizes input images resized at $256$x$256$ pixels. Our method performs competitively using only 1 labeled sample.}
   \label{table:state-of-the-art}
   \resizebox{1\textwidth}{!}{%
   \begin{tabular}{ l | c | c c c c | c c c c c | c c c c }
     \toprule
          \multirow{3}{*}{Methods} & \multirow{3}{*}{\shortstack{Number of Labeled \\Training Samples}} & \multicolumn{4}{c|}{\textbf{Chest}} & \multicolumn{5}{c|}{\textbf{Cephalometric}} & \multicolumn{4}{c}{\textbf{Hand}} \\
      & & MRE $\downarrow$ & \multicolumn{3}{c|}{SDR(\%) $\uparrow$} & MRE $\downarrow$ & \multicolumn{4}{c|}{SDR(\%) $\uparrow$} & MRE $\downarrow$ & \multicolumn{3}{c}{SDR(\%) $\uparrow$}  \\
     \cline{4-6} \cline{8-11} \cline{13-15}
     & & (px) & 3px & 6px & 9px & (mm) & 2mm & 2.5mm & 3mm & 4mm & (mm) & 2mm & 4mm & 10mm \\
     \toprule
     YOLO+    &   10 &    22.55 & 32.11 & 63.82 & 76.42 & 15.05 & 18.23 & 26.50 & 39.62 & 53.49 & 52.43 & 11.59 & 19.52 & 24.14 \\
     YOLO+    &   5 &    23.13 & 23.57 & 50.40 & 67.07 & 19.38 & 11.53 & 17.07 & 25.95 & 35.53 & 64.24 & 15.15 & 24.91 & 31.79 \\
     YOLO+    &   1 &    84.37 & 4.07 & 13.41 & 19.51 & 53.80 & 7.35 & 11.22 & 17.26 & 23.39 & 61.36 & 8.26 & 17.16 & 31.47     \\
     \midrule
     Our method   &   1 &    14.99 & 19.92 & 46.34 & 64.63 & 15.71 & 17.31 & 27.14 & 33.24 & 45.14 & 28.75 & 27.87 & 46.44 & 58.75      \\
     \bottomrule

   \end{tabular}%
   }
 \end{table*}

\begin{table*}[hbpt]
    \centering
    \caption{Landmark detection test performance of our Gaussian-heatmaps method, pre-trained on the Hand dataset and fine-tuned on the Chest / Cephalometric datasets, across varying numbers of labeled training images.}
    \label{tab:different_pretraining_datasets}
    {\small {
    \resizebox{1\textwidth}{!}{%
      \begin{tabular}{ c | c c c c | c c c c c }
        \toprule
        \multirow{3}{*}{\shortstack{Number of Labeled \\Training Samples}} &
        \multicolumn{4}{c|}{\textbf{Hand pre-training $\rightarrow$ Chest fine-tuning}} &
        \multicolumn{5}{c}{\textbf{Hand pre-training $\rightarrow$ Cephalometric fine-tuning}} \\
        
         & MRE$\pm$STD $\downarrow$ & \multicolumn{3}{c|}{SDR(\%)$\pm$STD $\uparrow$}
         & MRE$\pm$STD $\downarrow$ & \multicolumn{4}{c}{SDR(\%)$\pm$STD $\uparrow$} \\
        \cline{3-5} \cline{7-10}
               & (px) & 3px & 6px & 9px 
               & (mm) & 2mm & 2.5mm & 3mm & 4mm \\
        \toprule
            $1$        & $23.79\pm0.00$ & $14.23\pm0.00$ & $35.37\pm0.00$ & $52.03\pm0.00$ 
                        & $9.56\pm0.28$ & $30.45\pm1.73$ & $38.60\pm1.78$ & $47.75\pm2.00$ & $60.21\pm1.73$ \\
   
            $5$        & $5.86\pm0.00$ & $32.93\pm0.00$ & $65.04\pm0.00$ & $82.93\pm0.00$ 
                        & $5.42\pm0.00$ & $55.07\pm0.00$ & $64.55\pm0.00$ & $72.44\pm0.00$ & $79.71\pm0.00$ \\
                        
            $10$        & $5.08\pm0.00$ & $34.55\pm0.00$ & $70.73\pm0.00$ & $85.37\pm0.00$ 
                        & $2.78\pm0.00$ & $63.52\pm0.30$ & $74.20\pm0.62$ & $82.50\pm0.35$ & $89.97\pm0.35$ \\

            $25$        & $4.39\pm0.00$ & $38.62\pm0.00$ & $77.24\pm0.00$ & $91.46\pm0.00$ 
                        & $1.70\pm0.00$ & $74.03\pm0.01$ & $83.97\pm0.05$ & $90.95\pm0.04$ & $96.26\pm0.03$ \\

            $50$        & $3.99\pm0.00$ & $41.87\pm0.00$ & $82.52\pm0.00$ & $92.68\pm0.00$ 
                        & $1.50\pm0.01$ & $79.90\pm0.02$ & $87.85\pm0.02$ & $93.12\pm0.00$ & $97.37\pm0.00$ \\

        \bottomrule
      \end{tabular} }}
    }
\end{table*}

Finally, to contextualize our results, we compare our method to YOLO 
 \cite{zhu2021you}, a state-of-the-art universal anatomical landmark detection model. YOLO utilizes mixed dataset training and combines a universal U-Net for local feature learning with a global network for landmark disambiguation. As Table \ref{table:state-of-the-art} shows, our method performs comparably or better than YOLO variants across all three datasets. Despite YOLO's mixed dataset training (denoted by '$+$'), our approach achieves competitive results with just one labeled sample, even with $5$ and $10$ samples.

\paragraph{Ablation study on the downstream task setting.}
We perform an ablation study comparing the performance of our downstream task framework, with the one introduced in \cite{DBLP:conf/cvpr/McCouatV22} (second-last entry in our results Tables), mainly differing in the usage of a single annotated pixel to generate unrestricted contour-hugging heatmaps and a negative log-likelihood loss. 
For the Hand dataset, which had more extensive pre-training data, our approach consistently outperforms the evaluated alternative. However, the contour-hugging method generally shows better SDR performance on the Chest and the Cephalometric datasets. Regarding MRE, our original framework performs better up to $10$ training samples, showing lower performance with more annotated data. 
These findings suggest that the effectiveness of each downstream setting depends on pre-training data volume and quality, as well as dataset-specific characteristics. 

\subsection{Impact of different pre-training datasets} 
In this section, we investigate whether a different in-domain dataset can enhance performance when both available images and annotations are limited. We select the best DDPM model pre-trained on the larger Hand dataset and fine-tune it on the smaller Chest and Cephalometric datasets, following the same protocol as previous experiments with Gaussian heatmaps. Table \ref{tab:different_pretraining_datasets} reports the results.

Our experiments show the effectiveness of using a larger in-domain dataset (Hand) for pre-training in scenarios with limited target task data. For the Chest landmark detection task, the Hand pre-trained model outperforms the Chest pre-trained model in most scenarios (5 to 50 labeled images). Even with just one labeled sample, it achieves comparable performance to the Chest pre-trained model.
For the Cephalometric landmark detection task, while the Hand pre-trained model doesn't surpass the Cephalometric pre-trained model, it achieves competitive results across all training sample quantities. The small performance gap indicates that Hand pre-training still transfers useful features to the Cephalometric task.
Notably, our approach consistently outperforms other pre-training methods (ImageNet, MoCoV3, SimCLRV2, DINO) and frameworks (YOLO) for both tasks, regardless of the number of labeled training samples. This underscores the value of in-domain pre-training using our approach, even when the pre-training dataset differs from the target task.

Our goal is to simulate real-world scenarios with limited data and label availability by utilizing images from the same dataset as the downstream task. We focused on label-efficient pre-training while minimizing computational resource usage, given the substantial resources and time required for training diffusion models on extensive unlabeled datasets. While larger unlabeled datasets might enhance pre-training outcomes, our key finding is that using a limited unannotated dataset can still surpass self-supervised and supervised state-of-the-art pre-training techniques like ImageNet (which has one million images) when image and annotation availability is constrained.

\section{Conclusions}
\label{conclusion}
Our research introduces a novel application of DDPMs for anatomical landmark detection in X-ray images, addressing the persistent challenge of limited annotated data in medical image analysis. This work bridges the gap between the established potential of diffusion models in medical imaging and the critical need for label-efficient methods in landmark localization.

We demonstrate that our DDPM-based self-supervised pre-training method consistently outperforms supervised ImageNet pre-training, traditional state-of-the-art self-supervised approaches (MoCoV3, SimCLRV2, DINO), and the YOLO framework in few-shot learning scenarios across Chest, Cephalometric, and Hand X-ray datasets. The performance gains are particularly significant in low-data regimes, with substantial MRE reductions and SDR improvements. Notably, our method maintains its effectiveness when pre-trained on one in-domain dataset and fine-tuned on smaller, distinct datasets, simulating real-world scenarios with limited data availability. This robustness suggests broad applicability across various medical imaging tasks and datasets.

Our work not only addresses the critical issue of data scarcity but also contributes to the ongoing discussion on effective pre-training strategies in medical image analysis. By showing the efficacy of DDPMs in this context, we aims to open new avenues for research into efficient and robust landmark detection methods, potentially accelerating progress in automated diagnosis systems.

%%%%%%%%% REFERENCES
{\small
\bibliographystyle{ieee_fullname}
\bibliography{egbib}

\begin{thebibliography}{10}\itemsep=-1pt

\bibitem{Alzubaidi2021}
L. Alzubaidi, M. Al-Amidie, A. Al-Asadi, A.~J. Humaidi, O. Al-Shamma, M.~A. Fadhel, J. Zhang, J. Santamaría, and Y. Duan.
\newblock Novel transfer learning approach for medical imaging with limited labeled data.
\newblock {\em Cancers}, 13(7):1590, 2021.

\bibitem{DBLP:conf/iclr/BaranchukVRKB22}
Dmitry Baranchuk, Andrey Voynov, Ivan Rubachev, Valentin Khrulkov, and Artem Babenko.
\newblock Label-efficient semantic segmentation with diffusion models.
\newblock In {\em The Tenth International Conference on Learning Representations, {ICLR} 2022, Virtual Event, April 25-29, 2022}. OpenReview.net, 2022.

\bibitem{DBLP:journals/tmi/CandemirJPMSXKATM14}
Sema Candemir, Stefan Jaeger, Kannappan Palaniappan, Jonathan~P. Musco, Rahul~K. Singh, Zhiyun Xue, Alexandros Karargyris, Sameer~K. Antani, George~R. Thoma, and Clement~J. McDonald.
\newblock Lung segmentation in chest radiographs using anatomical atlases with nonrigid registration.
\newblock {\em {IEEE} Trans. Medical Imaging}, 33(2):577--590, 2014.

\bibitem{DBLP:conf/iccv/CaronTMJMBJ21}
Mathilde Caron, Hugo Touvron, Ishan Misra, Herv{\'{e}} J{\'{e}}gou, Julien Mairal, Piotr Bojanowski, and Armand Joulin.
\newblock Emerging properties in self-supervised vision transformers.
\newblock In {\em 2021 {IEEE/CVF} International Conference on Computer Vision, {ICCV} 2021, Montreal, QC, Canada, October 10-17, 2021}, pages 9630--9640. {IEEE}, 2021.

\bibitem{DBLP:conf/nips/ChenKSNH20}
Ting Chen, Simon Kornblith, Kevin Swersky, Mohammad Norouzi, and Geoffrey~E. Hinton.
\newblock Big self-supervised models are strong semi-supervised learners.
\newblock In Hugo Larochelle, Marc'Aurelio Ranzato, Raia Hadsell, Maria{-}Florina Balcan, and Hsuan{-}Tien Lin, editors, {\em Advances in Neural Information Processing Systems 33: Annual Conference on Neural Information Processing Systems 2020, NeurIPS 2020, December 6-12, 2020, virtual}, 2020.

\bibitem{DBLP:conf/iccv/ChenXH21}
Xinlei Chen, Saining Xie, and Kaiming He.
\newblock An empirical study of training self-supervised vision transformers.
\newblock In {\em 2021 {IEEE/CVF} International Conference on Computer Vision, {ICCV} 2021, Montreal, QC, Canada, October 10-17, 2021}, pages 9620--9629. {IEEE}, 2021.

\bibitem{DBLP:journals/corr/abs-2105-05233}
Prafulla Dhariwal and Alexander~Quinn Nichol.
\newblock Diffusion models beat gans on image synthesis.
\newblock In Marc'Aurelio Ranzato, Alina Beygelzimer, Yann~N. Dauphin, Percy Liang, and Jennifer~Wortman Vaughan, editors, {\em Advances in Neural Information Processing Systems 34: Annual Conference on Neural Information Processing Systems 2021, NeurIPS 2021, December 6-14, 2021, virtual}, pages 8780--8794, 2021.

\bibitem{edwards2021deepnavnet}
C.~A. Edwards, A. Goyal, A.~E. Rusheen, A.~Z. Kouzani, and K.~H. Lee.
\newblock Deepnavnet: Automated landmark localization for neuronavigation.
\newblock {\em Frontiers in Neuroscience}, 15:670287, 2021.

\bibitem{DBLP:conf/miccai/ElkhillLFP22}
Connor Elkhill, Scott LeBeau, Brooke French, and Antonio~R. Porras.
\newblock Graph convolutional network with probabilistic spatial regression: Application to craniofacial landmark detection from 3d photogrammetry.
\newblock In Linwei Wang, Qi Dou, P.~Thomas Fletcher, Stefanie Speidel, and Shuo Li, editors, {\em Medical Image Computing and Computer Assisted Intervention - {MICCAI} 2022 - 25th International Conference, Singapore, September 18-22, 2022, Proceedings, Part {III}}, volume 13433 of {\em Lecture Notes in Computer Science}, pages 574--583. Springer, 2022.

\bibitem{DBLP:journals/cmig/GertychZSPH07}
Arkadiusz Gertych, Aifeng Zhang, James~W. Sayre, Sylwia Pospiech{-}Kurkowska, and H.~K. Huang.
\newblock Bone age assessment of children using a digital hand atlas.
\newblock {\em Comput. Medical Imaging Graph.}, 31(4-5):322--331, 2007.

\bibitem{pyssl2023giakoumoglou}
Nikolaos Giakoumoglou and Paschalis Giakoumoglou.
\newblock Pyssl: A pytorch implementation of self-supervised learning (ssl) methods.
\newblock \url{https://github.com/giakou4/pyssl}, 2023.

\bibitem{DBLP:journals/corr/abs-2006-11239}
Jonathan Ho, Ajay Jain, and Pieter Abbeel.
\newblock Denoising diffusion probabilistic models.
\newblock In Hugo Larochelle, Marc'Aurelio Ranzato, Raia Hadsell, Maria{-}Florina Balcan, and Hsuan{-}Tien Lin, editors, {\em Advances in Neural Information Processing Systems 33: Annual Conference on Neural Information Processing Systems 2020, NeurIPS 2020, December 6-12, 2020, virtual}, 2020.

\bibitem{Iakubovskii:2019}
Pavel Iakubovskii.
\newblock Segmentation models pytorch.
\newblock \url{https://github.com/qubvel/segmentation_models.pytorch}, 2019.

\bibitem{ibragimov2017landmark}
Bulat Ibragimov and Toma{\v{z}} Vrtovec.
\newblock Landmark-based statistical shape representations.
\newblock In {\em Statistical Shape and Deformation Analysis}, pages 89--113. Elsevier, 2017.

\bibitem{DBLP:conf/miccai/JiangLWTLL22}
Yankai Jiang, Yiming Li, Xinyue Wang, Yubo Tao, Jun Lin, and Hai Lin.
\newblock Cephalformer: Incorporating global structure constraint into visual features for general cephalometric landmark detection.
\newblock In Linwei Wang, Qi Dou, P.~Thomas Fletcher, Stefanie Speidel, and Shuo Li, editors, {\em Medical Image Computing and Computer Assisted Intervention - {MICCAI} 2022 - 25th International Conference, Singapore, September 18-22, 2022, Proceedings, Part {III}}, volume 13433 of {\em Lecture Notes in Computer Science}, pages 227--237. Springer, 2022.

\bibitem{kasturi2024anatomical}
Akhil Kasturi, Ali Vosoughi, Nathan Hadjiyski, Larry Stockmaster, William~J Sehnert, and Axel Wism{\"u}ller.
\newblock Anatomical landmark detection in chest x-ray images using transformer-based networks.
\newblock In {\em Medical Imaging 2024: Computer-Aided Diagnosis}, volume 12927, pages 647--660. SPIE, 2024.

\bibitem{DBLP:journals/mia/KazerouniAHAFHM23}
Amirhossein Kazerouni, Ehsan~Khodapanah Aghdam, Moein Heidari, Reza Azad, Mohsen Fayyaz, Ilker Hacihaliloglu, and Dorit Merhof.
\newblock Diffusion models in medical imaging: {A} comprehensive survey.
\newblock {\em Medical Image Anal.}, 88:102846, 2023.

\bibitem{DBLP:journals/bmcmi/KimCSJMG22}
Hee~E. Kim, Alejandro Cosa{-}Linan, Nandhini Santhanam, Mahboubeh Jannesari, M{\'{a}}t{\'{e}}~E. Maros, and Thomas Ganslandt.
\newblock Transfer learning for medical image classification: a literature review.
\newblock {\em {BMC} Medical Imaging}, 22(1):69, 2022.

\bibitem{DBLP:journals/ijon/LeeCS22}
Minkyung Lee, Minyoung Chung, and Yeong{-}Gil Shin.
\newblock Cephalometric landmark detection via global and local encoders and patch-wise attentions.
\newblock {\em Neurocomputing}, 470:182--189, 2022.

\bibitem{DBLP:conf/cvpr/McCouatV22}
James McCouat and Irina Voiculescu.
\newblock Contour-hugging heatmaps for landmark detection.
\newblock In {\em {IEEE/CVF} Conference on Computer Vision and Pattern Recognition, {CVPR} 2022, New Orleans, LA, USA, June 18-24, 2022}, pages 20565--20573. {IEEE}, 2022.

\bibitem{DBLP:conf/isbi/McCouatVG21}
James McCouat, Irina Voiculescu, and Si{\^{o}}n Glyn{-}Jones.
\newblock Automatically diagnosing {HIP} conditions from x-rays using landmark detection.
\newblock In {\em 18th {IEEE} International Symposium on Biomedical Imaging, {ISBI} 2021, Nice, France, April 13-16, 2021}, pages 179--182. {IEEE}, 2021.

\bibitem{Meijering2020}
Erik Meijering.
\newblock Deep learning in bioimaging.
\newblock {\em Computational and Structural Biotechnology Journal}, 18:2301--2313, 2020.

\bibitem{DBLP:journals/corr/abs-2407-05412}
Juzheng Miao, Cheng Chen, Keli Zhang, Jie Chuai, Quanzheng Li, and Pheng{-}Ann Heng.
\newblock {FM-OSD:} foundation model-enabled one-shot detection of anatomical landmarks.
\newblock {\em CoRR}, abs/2407.05412, 2024.

\bibitem{DBLP:journals/corr/abs-2212-07501}
Gustav M{\"{u}}ller{-}Franzes, Jan~Moritz Niehues, Firas Khader, Soroosh~Tayebi Arasteh, Christoph Haarburger, Christiane Kuhl, Tianci Wang, Tianyu Han, Sven Nebelung, Jakob~Nikolas Kather, and Daniel Truhn.
\newblock Diffusion probabilistic models beat gans on medical images.
\newblock {\em CoRR}, abs/2212.07501, 2022.

\bibitem{DBLP:journals/mia/PayerSBU19}
Christian Payer, Darko Stern, Horst Bischof, and Martin Urschler.
\newblock Integrating spatial configuration into heatmap regression based cnns for landmark localization.
\newblock {\em Medical Image Anal.}, 54:207--219, 2019.

\bibitem{DBLP:conf/miccai/PinayaTDCFNOC22}
Walter H.~L. Pinaya, Petru{-}Daniel Tudosiu, Jessica Dafflon, Pedro F.~Da Costa, Virginia Fernandez, Parashkev Nachev, S{\'{e}}bastien Ourselin, and M.~Jorge Cardoso.
\newblock Brain imaging generation with latent diffusion models.
\newblock In Anirban Mukhopadhyay, Ilkay {\"{O}}ks{\"{u}}z, Sandy Engelhardt, Dajiang Zhu, and Yixuan Yuan, editors, {\em Deep Generative Models - Second {MICCAI} Workshop, {DGM4MICCAI} 2022, Held in Conjunction with {MICCAI} 2022, Singapore, September 22, 2022, Proceedings}, volume 13609 of {\em Lecture Notes in Computer Science}, pages 117--126. Springer, 2022.

\bibitem{DBLP:conf/cvpr/QuanYLZ22}
Quan Quan, Qingsong Yao, Jun Li, and S.~Kevin Zhou.
\newblock Which images to label for few-shot medical landmark detection?
\newblock In {\em {IEEE/CVF} Conference on Computer Vision and Pattern Recognition, {CVPR} 2022, New Orleans, LA, USA, June 18-24, 2022}, pages 20574--20584. {IEEE}, 2022.

\bibitem{DBLP:conf/miccai/RousseauACMMA23}
J{\'{e}}r{\'{e}}my Rousseau, Christian Alaka, Emma Covili, Hippolyte Mayard, Laura Misrachi, and Willy Au.
\newblock Pre-training with diffusion models for dental radiography segmentation.
\newblock In Anirban Mukhopadhyay, Ilkay {\"{O}}ks{\"{u}}z, Sandy Engelhardt, Dajiang Zhu, and Yixuan Yuan, editors, {\em Deep Generative Models - Third {MICCAI} Workshop, {DGM4MICCAI} 2023, Held in Conjunction with {MICCAI} 2023, Vancouver, BC, Canada, October 8, 2023, Proceedings}, volume 14533 of {\em Lecture Notes in Computer Science}, pages 174--182. Springer, 2023.

\bibitem{DBLP:journals/ijmir/SuganyadeviS022}
S. Suganyadevi, V. Seethalakshmi, and K. Balasamy.
\newblock A review on deep learning in medical image analysis.
\newblock {\em Int. J. Multim. Inf. Retr.}, 11(1):19--38, 2022.

\bibitem{DBLP:conf/iccvw/TiulpinMS19}
Aleksei Tiulpin, Iaroslav Melekhov, and Simo Saarakkala.
\newblock {KNEEL:} knee anatomical landmark localization using hourglass networks.
\newblock In {\em 2019 {IEEE/CVF} International Conference on Computer Vision Workshops, {ICCV} Workshops 2019, Seoul, Korea (South), October 27-28, 2019}, pages 352--361. {IEEE}, 2019.

\bibitem{touijer2023food}
Larbi Touijer, Vito~Paolo Pastore, and Francesca Odone.
\newblock Food image classification: The benefit of in-domain transfer learning.
\newblock In {\em International Conference on Image Analysis and Processing}, pages 259--269. Springer, 2023.

\bibitem{DBLP:journals/bmcmi/VanBerloHW24}
Blake VanBerlo, Jesse Hoey, and Alexander Wong.
\newblock A survey of the impact of self-supervised pretraining for diagnostic tasks in medical x-ray, ct, mri, and ultrasound.
\newblock {\em {BMC} Medical Imaging}, 24(1):79, 2024.

\bibitem{divia2024indomain}
Roberto~Di Via, Matteo Santacesaria, Francesca Odone, and Vito~Paolo Pastore.
\newblock Is in-domain data beneficial in transfer learning for landmarks detection in x-ray images?
\newblock {\em CoRR}, abs/2403.01470, 2024.

\bibitem{DBLP:conf/miccai/ViriyasaranonMC23}
Thanaporn Viriyasaranon, Serie Ma, and Jang~Hwan Choi.
\newblock Anatomical landmark detection using a multiresolution learning approach with a hybrid transformer-cnn model.
\newblock In Hayit Greenspan, Anant Madabhushi, Parvin Mousavi, Septimiu Salcudean, James Duncan, Tanveer~F. Syeda{-}Mahmood, and Russell~H. Taylor, editors, {\em Medical Image Computing and Computer Assisted Intervention - {MICCAI} 2023 - 26th International Conference, Vancouver, BC, Canada, October 8-12, 2023, Proceedings, Part {VI}}, volume 14225 of {\em Lecture Notes in Computer Science}, pages 433--443. Springer, 2023.

\bibitem{DBLP:journals/mia/WangHLLCSLIVRFC16}
Ching{-}Wei Wang, Cheng{-}Ta Huang, Jia{-}Hong Lee, Chung{-}Hsing Li, Sheng{-}Wei Chang, Ming{-}Jhih Siao, Tat{-}Ming Lai, Bulat Ibragimov, Tomaz Vrtovec, Olaf Ronneberger, Philipp Fischer, Timothy~F. Cootes, and Claudia Lindner.
\newblock A benchmark for comparison of dental radiography analysis algorithms.
\newblock {\em Medical Image Anal.}, 31:63--76, 2016.

\bibitem{DBLP:journals/corr/abs-2206-02087}
Zhiwei Wang, Jinxin Lv, Yunqiao Yang, Yuanhuai Liang, Yi Lin, Qiang Li, Xin Li, and Xin Yang.
\newblock Accurate scoliosis vertebral landmark localization on x-ray images via shape-constrained multi-stage cascaded cnns.
\newblock {\em CoRR}, abs/2206.02087, 2022.

\bibitem{DBLP:conf/cvpr/WyattLSW22}
Julian Wyatt, Adam Leach, Sebastian~M. Schmon, and Chris~G. Willcocks.
\newblock Anoddpm: Anomaly detection with denoising diffusion probabilistic models using simplex noise.
\newblock In {\em {IEEE/CVF} Conference on Computer Vision and Pattern Recognition Workshops, {CVPR} Workshops 2022, New Orleans, LA, USA, June 19-20, 2022}, pages 649--655. {IEEE}, 2022.

\bibitem{DBLP:conf/eccv/XieR18}
Yiting Xie and David Richmond.
\newblock Pre-training on grayscale imagenet improves medical image classification.
\newblock In Laura Leal{-}Taix{\'{e}} and Stefan Roth, editors, {\em Computer Vision - {ECCV} 2018 Workshops - Munich, Germany, September 8-14, 2018, Proceedings, Part {VI}}, volume 11134 of {\em Lecture Notes in Computer Science}, pages 476--484. Springer, 2018.

\bibitem{DBLP:conf/miccai/YaoHHZ20}
Qingsong Yao, Zecheng He, Hu Han, and S.~Kevin Zhou.
\newblock Miss the point: Targeted adversarial attack on multiple landmark detection.
\newblock In Anne~L. Martel, Purang Abolmaesumi, Danail Stoyanov, Diana Mateus, Maria~A. Zuluaga, S.~Kevin Zhou, Daniel Racoceanu, and Leo Joskowicz, editors, {\em Medical Image Computing and Computer Assisted Intervention - {MICCAI} 2020 - 23rd International Conference, Lima, Peru, October 4-8, 2020, Proceedings, Part {IV}}, volume 12264 of {\em Lecture Notes in Computer Science}, pages 692--702. Springer, 2020.

\bibitem{DBLP:conf/miccai/YaoQXZ21}
Qingsong Yao, Quan Quan, Li Xiao, and S.~Kevin Zhou.
\newblock One-shot medical landmark detection.
\newblock In Marleen de Bruijne, Philippe~C. Cattin, St{\'{e}}phane Cotin, Nicolas Padoy, Stefanie Speidel, Yefeng Zheng, and Caroline Essert, editors, {\em Medical Image Computing and Computer Assisted Intervention - {MICCAI} 2021 - 24th International Conference, Strasbourg, France, September 27 - October 1, 2021, Proceedings, Part {II}}, volume 12902 of {\em Lecture Notes in Computer Science}, pages 177--188. Springer, 2021.

\bibitem{DBLP:conf/eccv/YinGWYW22}
Zihao Yin, Ping Gong, Chunyu Wang, Yizhou Yu, and Yizhou Wang.
\newblock One-shot medical landmark localization by edge-guided transform and noisy landmark refinement.
\newblock In Shai Avidan, Gabriel~J. Brostow, Moustapha Ciss{\'{e}}, Giovanni~Maria Farinella, and Tal Hassner, editors, {\em Computer Vision - {ECCV} 2022 - 17th European Conference, Tel Aviv, Israel, October 23-27, 2022, Proceedings, Part {XXI}}, volume 13681 of {\em Lecture Notes in Computer Science}, pages 473--489. Springer, 2022.

\bibitem{DBLP:conf/miccai/ZhuQYLZ23}
Heqin Zhu, Quan Quan, Qingsong Yao, Zaiyi Liu, and S.~Kevin Zhou.
\newblock {UOD:} universal one-shot detection of anatomical landmarks.
\newblock In Hayit Greenspan, Anant Madabhushi, Parvin Mousavi, Septimiu Salcudean, James Duncan, Tanveer~F. Syeda{-}Mahmood, and Russell~H. Taylor, editors, {\em Medical Image Computing and Computer Assisted Intervention - {MICCAI} 2023 - 26th International Conference, Vancouver, BC, Canada, October 8-12, 2023, Proceedings, Part {I}}, volume 14220 of {\em Lecture Notes in Computer Science}, pages 24--34. Springer, 2023.

\bibitem{Zhu2022}
H Zhu, Q Yao, L Xiao, and SK Zhou.
\newblock Learning to localize cross-anatomy landmarks in x-ray images with a universal model.
\newblock {\em BME Front.}, 2022:9765095, Jun 2022.

\bibitem{zhu2021you}
Heqin Zhu, Qingsong Yao, Li Xiao, and S.~Kevin Zhou.
\newblock You only learn once: Universal anatomical landmark detection.
\newblock In Marleen de Bruijne, Philippe~C. Cattin, St{\'{e}}phane Cotin, Nicolas Padoy, Stefanie Speidel, Yefeng Zheng, and Caroline Essert, editors, {\em Medical Image Computing and Computer Assisted Intervention - {MICCAI} 2021 - 24th International Conference, Strasbourg, France, September 27 - October 1, 2021, Proceedings, Part {V}}, volume 12905 of {\em Lecture Notes in Computer Science}, pages 85--95. Springer, 2021.

\end{thebibliography}
}

\end{document}